\documentclass[journal]{IEEEtran}

\ifCLASSINFOpdf
\else
\fi
\usepackage{times}
\usepackage{xcolor}
\usepackage{soul}
\usepackage[utf8]{inputenc}
\usepackage[small]{caption}
\usepackage{multirow}
\usepackage{graphicx}  
\usepackage{helvet}
\usepackage{courier}
\usepackage{url}
\usepackage{color}
\usepackage{amsmath}
\usepackage{amssymb}
\usepackage{algorithm}
\usepackage{algorithmic}
\newtheorem{myDef}{Definition}

\def\ie{{\em i.e.}}
\def\eg{{\em e.g.}}

\begin{document}
%
\title{Multi-Drone based Single Object Tracking with Agent Sharing Network}
%
\author{Pengfei Zhu, Jiayu Zheng, Dawei Du, Longyin Wen, Yiming Sun, Qinghua Hu
\IEEEcompsocitemizethanks{
\IEEEcompsocthanksitem Pengfei Zhu, Jiayu Zheng, Yiming Sun and Qinghua Hu are with the College of Intelligence and Computing, Tianjin University, Tianjin, China (e-mail: \{zhupengfei,zhengjiayu,sunyiming1895, huqinghua\}@tju.edu.cn).
\IEEEcompsocthanksitem Longyin Wen is with the JD Finance America Corporation, Mountain View, CA, USA (e-mail: longyin.wen@jd.com).
\IEEEcompsocthanksitem Dawei Du is with the Computer Science Department, University at Albany, State University of New York, Albany, NY, USA (e-mail: ddu@albany.edu).
\IEEEcompsocthanksitem This work was supported by the National Natural Science Foundation of China under Grants 61502332, 61876127 and 61732011, Natural Science Foundation of Tianjin under Grant 17JCZDJC30800, and Key Scientific and Technological Support Projects of Tianjin Key R\&D Program under Grant 18YFZCGX00390 and Grant 18YFZCGX00680.}
}

\markboth{}%
{Shell \MakeLowercase{\textit{et al.}}: Bare Demo of IEEEtran.cls for IEEE Journals}

\maketitle

\begin{abstract}
Drone equipped with cameras can dynamically track the target in the air from a broader view compared with static cameras or moving sensors over the ground. However, it is still challenging to accurately track the target using a single drone due to several factors such as appearance variations and severe occlusions. In this paper, we collect a new \textbf{M}ulti-\textbf{D}rone single \textbf{O}bject \textbf{T}racking (\textbf{MDOT}) dataset that consists of $92$ groups of video clips with $113,918$ high resolution frames taken by two drones and $63$ groups of video clips with $145,875$ high resolution frames taken by three drones. Besides, two evaluation metrics are specially designed for multi-drone single object tracking, \ie, automatic fusion score (AFS) and ideal fusion score (IFS).
Moreover, an agent sharing network (ASNet) is proposed by self-supervised template sharing and view-aware fusion of the target from multiple drones, which can improve the tracking accuracy significantly compared with single drone tracking.
Extensive experiments on MDOT show that our ASNet significantly outperforms recent state-of-the-art trackers.
\end{abstract}


\section{Introduction}
Drone or general unmanned aerial vehicle (UAV) is widely used in our daily life.
Specifically, the application scenarios of drone based visual tracking cover live broadcast, military battlefield, criminal investigation, sports and entertainment \cite{uav123,visdrone2018,du2018unmanned}.
Compared with static cameras and handheld mobile devices, drones can dynamically move and cover a wide ground area, which is very suitable to track fast-moving targets.

\begin{figure}[t]
	\centering
	\includegraphics[width=0.95\linewidth]{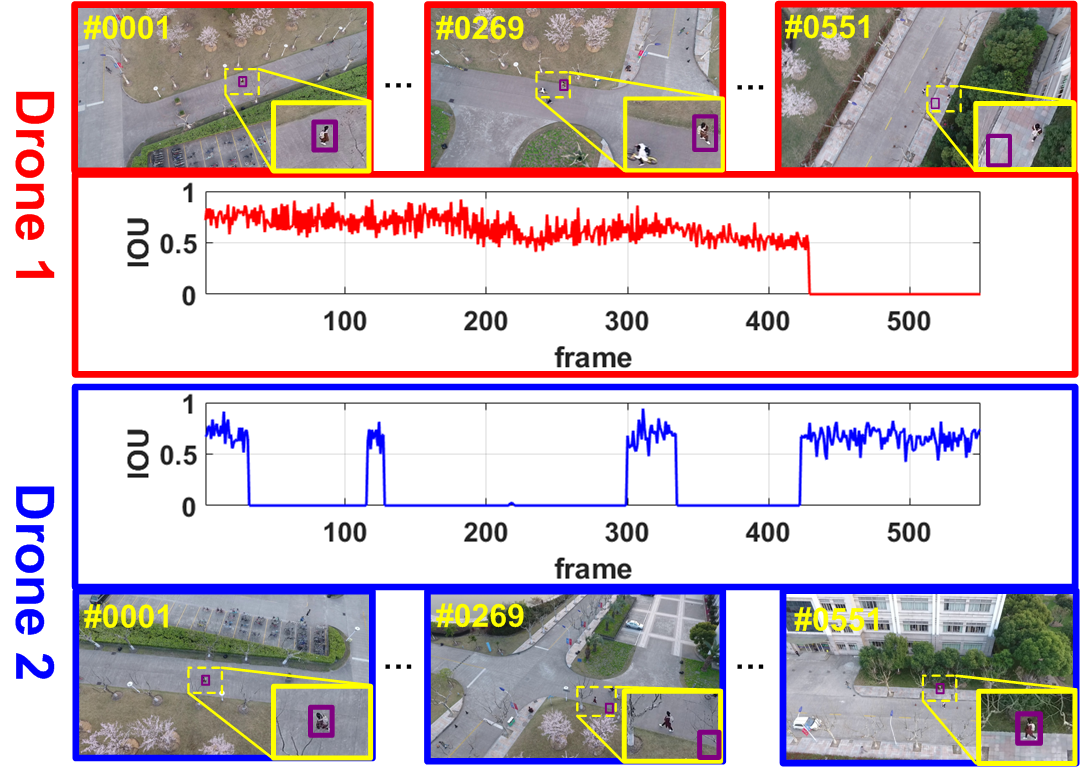}
	\caption{An illustration of tracking the same pedestrian using two drones. The purple bounding boxes denote the tracking results. The second and third rows show the intersection over union (IOU) between the tracking result and groundtruth for the first and second drone, respectively. Each base tracker may fail in tracking only based on one drone.}
	\label{architecture}
\end{figure}

To perform robust tracking on drones, large-scale datasets with high-quality annotations play a critical role to promote the development of algorithms. Recently, several benchmark datasets have been collected with a single drone, including UAV123 \cite{uav123}, Campus \cite{robicquet2016learning}, VisDrone-2018 \cite{visdrone2018}, and UAVDT \cite{du2018unmanned}. As shown in Figure~\ref{architecture}, compared with static cameras that only collect data in a certain area, the drone can also dynamically track the target in the air from a broad view. However, it also brings additional challenges to visual tracking, including tiny targets, camera motion, high density distribution of targets, \textit{etc}.

To solve the above issue, incorporating multiple drones is an effective solution to improve the performance and robustness of object tracking to occlusion and appearance ambiguities. Thus, several algorithms focus on long-term tracking and re-identification based on multiple cameras in video surveillance \cite{cai2014exploring,narayan2017person,ristani2018features,chen2014novel,chen2017integrating,houssineau2016unified,ristani2016performance,chen2017equalized,lee2018online}.
In the past few years, a few multi-camera benchmark datasets have been constructed with overlapping or non-overlapping views of cameras \cite{berclaz2011multiple,zhang2015camera,ristani2016performance}.
Some datasets with fully overlapped views are constrained to short time intervals and controlled conditions \cite{berclaz2011multiple}.
These multi-camera datasets are specially collected for multi-object tracking or person re-identification across cameras.

Although many datasets are provided for visual tracking, they are built for either single drone tracking or multi-camera tracking.
However, there are few benchmark datasets for multi-drone visual tracking. In this paper, to combine the advantages of both drone based tracking and multi-camera tracking, we present a multi-drone single object tracking (MDOT) dataset.
MDOT consists of $92$ groups of video clips with $113,918$ high resolution frames taken by two drones and $63$ groups of video clips with $145,875$ high resolution frames taken by three drones.
In each group of video clips, the same target is tracked by multiple drones. Moreover, we annotate $10$ different types of attributes, including \textit{daylight}, \textit{night}, \textit{camera motion}, \textit{partial occlusion}, \textit{full occlusion}, \textit{out of view}, \textit{similar object}, \textit{viewpoint change}, \textit{illumination variation}, and \textit{low resolution}. To evaluate tracking algorithms in our dataset, we propose two new evaluation metrics, \ie, adaptive fusion score (AFS) and ideal fusion score (IFS). Specifically, AFS measures the performance of multi-drone tracker that fuses the tracking results using the online fusing strategy, while IFS is the ideal fusion performance when we assume that the multi-drone system can accurately select the tracking results of the drone with better performance. On the other hand, to use multi-drone complementarity, we propose an agent sharing network (ASNet), which shares templates across drones in a self-supervised manner and fuses the tracking results automatically for robust and accurate visual tracking. A re-detection strategy on drone based tracker is proposed to deal with drift of targets by enlarging the size of search region when the target is judged to satisfy the defined condition. Experiments on MDOT demonstrate that ASNet greatly outperforms $20$ recent state-of-the-art tracking algorithms.

\begin{figure}
	\centering
	\includegraphics[width=0.95\linewidth]{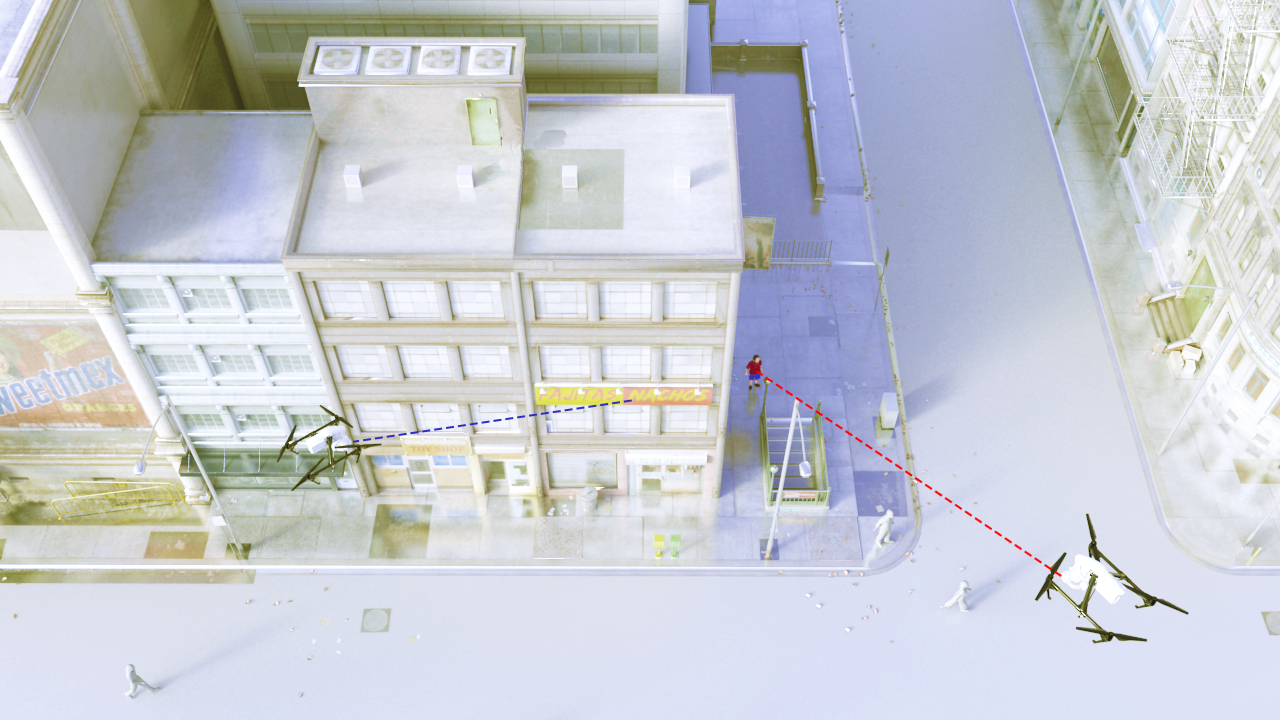}
	\caption{Illustration of data collection with two drones. The pedestrian on the edge of the building is tracked by two drones at the same time.}
	\label{camera}
\end{figure}

The contribution of this paper can be summarized as follows.
\begin{itemize}
  \item To the best of our knowledge, we propose the \textbf{first multi-drone single object tracking dataset (MDOT)} which consists of total $155$ groups of video clips with $259,793$ high resolution frames and rich annotations.
  \item Two new evaluation metrics are designed for multi-drone single object tracking, \ie, automatic fusion score (AFS) and ideal fusion score (IFS).
AFS evaluates the performance of multi-drone tracker that fuses the tracking results using the learned weights while IFS is the theoretically optimal fusion performance (upper bound) of a multi-drone tracker.
IFS is specially proposed to inspire researchers to design superior multi-drone tracker with more useful fusion strategies.
  \item The agent sharing network (ASNet) is proposed for perform multi-drone visual tracking, which effectively exploits multi-drone shared information in a view-aware manner.
  ASNet can be considered as a baseline tracker for the multi-drone based single object tracking task.
\end{itemize}
\section{Related Work}

\subsection{Single Object Tracking.}
In the field of visual tracking, single object tracking has achieved massive attention.
Generally, single object trackers can be categorized into generative models and discriminative ones.
Generative models search for the most similar area to the template of the previous frames, e..g, Kalman filtering and mean-shift \cite{comaniciu2000real}.
Discriminative models treat visual tracking as a binary classification task, which distinguish the target from the background, e.g., Struck \cite{hare2016struck} and TLD \cite{kalal2012tracking}.

Inspired by the success of deep learning in image classification and object detection, deep trackers have achieved superior performances. MDNet learns an end-to-end deep tracker upon convolutional neural network by a video-specific fully connected layer \cite{nam2016learning}. Siam-FC exploits a siamese network by learning the feature maps of both the target and the search region, and using a convolutional operation to obtain the response map \cite{bertinetto2016fully}. Recently, some successful techniques in object detection, \eg, region proposal network (RPN), are embedded into object tracking models. SiamRPN and its variants (DaSiamRPN \cite{zhu2018distractor}, SiamRPN++ \cite{li2018siamrpn++} and C-RPN \cite{fan2018siamese}), are proposed by using more powerful network structure as backbone and attractive blocks to learn feature maps with better representation abilities.
Discriminative correlation filter (DCF) can be learned very efficiently in the frequency domain via fast Fourier transform and therefore achieves very impressive tracking efficiency \cite{henriques2015high,ma2015long,danelljan2015learning,mueller2017context,kiani2017learning,lukezic2017discriminative,li2018learning,xu2019joint}.
To cope with the severe variations, discriminative feature representation, nonlinear kernel, scale estimation, spatial regularization, continuous convolution, spatial-temporal regularization are introduced to pursue a balance between accuracy and speed for correlation filter based trackers.
The performance of single object tracking is easily affected by severe appearance variations, occlusions, and out-of view cases, which could be solved by using multiple cameras.
\begin{figure*}[t]
\centering
  \includegraphics[width=0.95\linewidth]{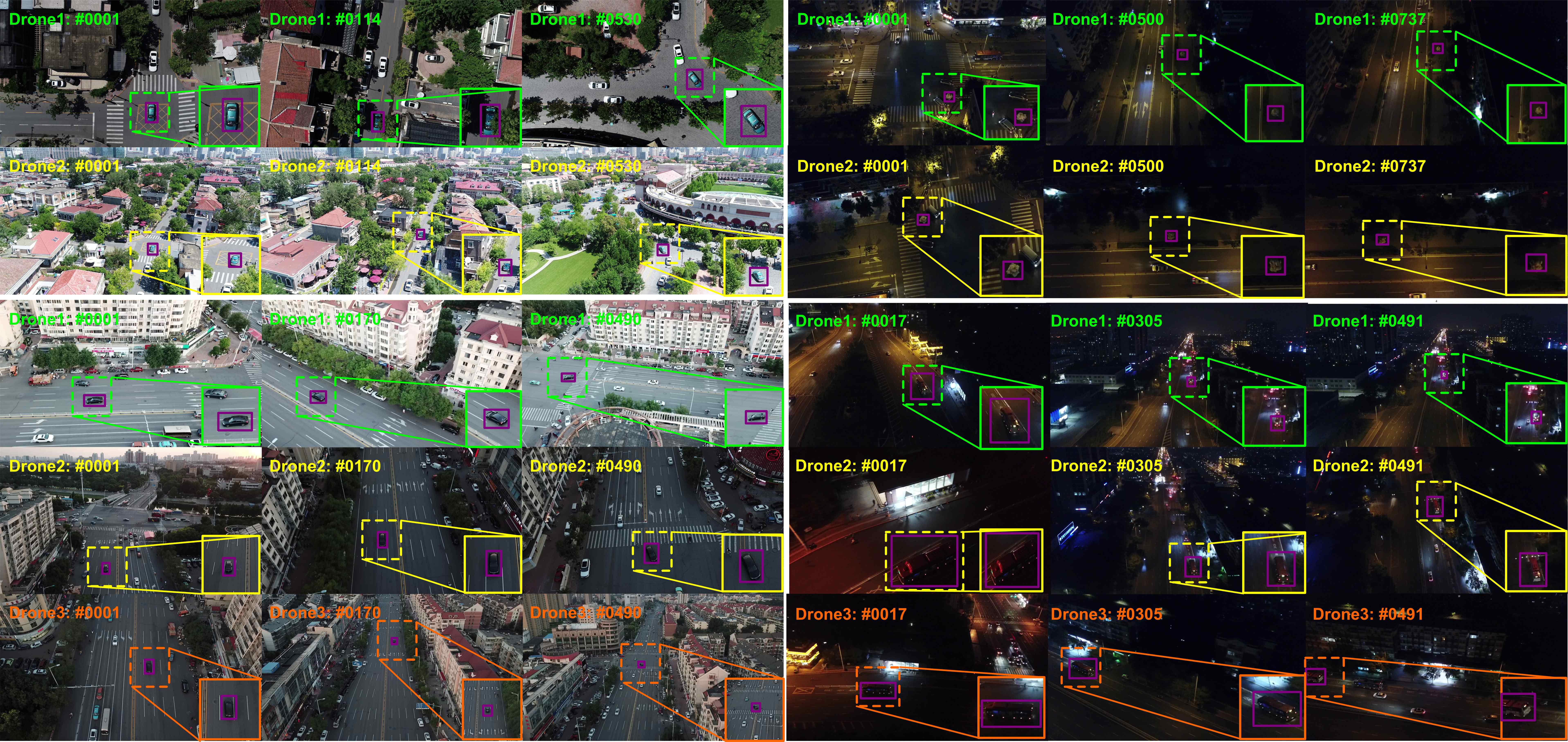}
\caption{Visual examples in our multi-drone single tracking (MDOT) dataset. The first two rows indicates the sequences from the Two-MDOT subset, while the last three rows indicates the sequences from the Three-MDOT subset. The purple bounding box denotes the ground-truth box.}
\label{dataset}
\end{figure*}

\begin{table*}[t]
	\centering
	\caption{Comparison of existing datasets for visual tracking ($1k=1,000, 1M=1,000k$).}
	\label{tab:comparison-dataset}
	 \footnotesize
	 \setlength{\tabcolsep}{15.0pt}
	\begin{tabular}{c|c|c|c|c|c}
		\hline
		Datasets                                               &Scenarios &Sequences   &Frames  & \# of Cameras  &Year \\
		\hline
		ALOV300 \cite{DBLP:journals/pami/SmeuldersCCCDS14}     &life      &$314$       &$151.6k$  & Single   &2014 \\
		OTB100 \cite{wu2015object}                             &life      &$100$       &$59.0k$   & Single   &2015 \\
		TC128 \cite{liang2015encoding}                         &life      &$128$       &$55.3k$   & Single   &2015 \\
		VOT2016 \cite{vot2016}                                 &life      &$60$        &$21.5k$   & Single   &2016 \\
		UAV123 \cite{uav123}                                   &drone     &$123$       &$110k$    & Single   &2016 \\
		NfS \cite{kiani2017need}                               &life      &$100$       &$383k$    & Single   &2017 \\
		POT210 \cite{DBLP:journals/corr/LiangWL17}             &planar objects &$210$  &$105.2k$  & Single   &2018 \\
		UAVDT \cite{du2018unmanned}                            &drone     &$100$       &$80k$     & Single   &2018 \\		
		VisDrone-SOT2018 \cite{visdrone2018}                   &drone     &$132$       &$106.4k$  & Single   &2018 \\
		LaSOT \cite{DBLP:conf/cvpr/FanLYCDYBXLL19}             &life      &$1400$      &$3.5M$    & Single   &2019 \\
		DukeMTMC \cite{ristani2016performance}                 &life      &$8$         &$2.9M$    & Multiple &2016 \\
		NLPR\_MCT \cite{chen2016equalised}                     &life      &$15$        &$89.5k$   & Multiple &2016 \\		
		\hline
		Two-MDOT                                               &drone     &$184$       &$113.9k$  & Multiple &2019 \\
		Three-MDOT                                             &drone     &$189$       &$145.9k$  & Multiple &2019 \\
		MDOT                                                   &drone     &$373$       &$259.8k$  & Multiple &2019 \\
		\hline
	\end{tabular}
\end{table*}

\subsection{Multi-Camera Tracking.}
Multi-camera tracking uses information of different views by estimating a common axis or subspace, or fusing multi-view information, to improve the robustness of trackers to occlusion, drift and other variations.
Most existing works focus on multi-object tracking, especially multi-person detection and tracking in overlapping views or across non-overlapping views \cite{narayan2017person,chen2017integrating,ristani2016performance,ristani2018features,lee2018online}.
The single-camera trajectories can be previously given or obtained by pedestrian detection and tracking.
As multiple cameras are static, the spatial relations between cameras are either explicitly mapped in 3D, learned by tracking
known identities, or obtained by comparing entry/exit rates across pairs of cameras.
The multi-camera information can be fused in different stages. Single-camera tracking is first performed in each camera to create trajectories of multiple targets, and then inter-camera tracking is carried out to associate the tracks \cite{lee2018online}. Trackers exploit optimization to maximize the coherence of observations for predicted identities. The spatial, temporal and appearance information of trajectories is used to construct an affinity matrix. The nodes are then partitioned into different identities by bipartite matching or maximal internal weights.

\subsection{Datasets}
The recent progresses in visual tracking rely on scarcely-available large scale tracking datasets with high-quality annotations to a great extent.
There have been significant growth of the volume and diversity of benchmark datasets for visual tracking, \eg, TC-128 \cite{liang2015encoding}, OTB-2015 \cite{wu2015object}, NUS-PRO \cite{NUS-PRO}, VOT2016 \cite{vot2016}, LaSOT \cite{fan2019lasot}, and TrackingNet \cite{Muller_2018_ECCV}).
Most datasets focus on single camera and single object tracking.
For drone based visual tracking, several benchmark datasets have been collected with a single drone, including UAV123 \cite{uav123}, Campus \cite{robicquet2016learning}, VisDrone-2018 \cite{visdrone2018}, and UAVDT \cite{du2018unmanned}.
UAV123 dataset contains a total of 123 video sequences and more than 110K frames for single object tracking \cite{uav123}.
Campus dataset includes 929.5k frames which contains various types of objects \cite{robicquet2016learning}.
VisDrone-2018 dataset consists of 263 video clips formed by 179,264 frames and 10,209 static images from 14 different cities for object detection and tracking in both images and videos \cite{visdrone2018}.
UAVDT dataset consists of 80,000 representative frames with bounding boxes as well as up to 14 kinds of attributes from 10 hours raw videos for object detection and tracking in videos \cite{du2018unmanned}.
To track and identify multiple objects across different cameras, a few multi-camera benchmark datasets are collected for multi-object tracking and person re-identification.
NLPR-MCT consists of four subsets with at most $255$ IDs with $3$ to $5$ non-overlapping cameras \cite{chen2016equalised}. DukeMTMC is the largest multi-camera multi-object tracking dataset that consists of videos of $2834$ IDs and $8$ cameras in the outdoor scene with both overlapping views and blind spots \cite{ristani2016performance}. All trajectories are manually annotated and identities are associated across cameras.
The existing datasets are collected either for single object tracking or multi-camera tracking using static sensors.
In this work, we collect a benchmark dataset using multiple drones, which is an effective supplement to the existing datasets.

\section{Multi-Drone Single Object Tracking Dataset}
In this section, we present the collected benchmark dataset (MDOT) and evaluation metrics for multi-drone single object tracking.

\subsection{Data Collection}
Our dataset is collected by multiple DJI PHANTOM 4Pro drones. Specifically, the drones are controlled by several professional human operators from different altitudes in various outdoor scenes (\eg, park, campus, square, and street), as illustrated in Figure \ref{camera}. In order to increase the targets' diversity of appearances and scales, the same target is tracked by multiple drones from different view-angles and different altitudes, ranging from $20$m to $100$m. The dataset has $155$ groups of video clips with $259,793$ high resolution frames in two sub-datasets. The two-drone based dataset (Two-MDOT) consists of $92$ groups of video clips with $113,918$ high resolution frames taken by two drones, while the three-drone based dataset (Three-MDOT) contains $63$ groups of video clips with $145,875$ high resolution frames taken by three drones.
Two-MDOT was collected in 2018 while Three-MDOT was collected in 2019. Hence, there is no overlap between Two-MDOT and Three-MDOT.
Besides, the dataset is divided into \textit{train} set ($37$ groups in Two-MDOT and $28$ groups of in Three-MDOT) and the \textit{test} set ($55$ groups in Two-MDOT and $35$ groups of in Three-MDOT).

As presented in Table \ref{tab:comparison-dataset}, most of the previous datasets are collected by one camera where the appearance of targets is not abundant.
Although NLPR\_MCT and DukeMTMC are used for evaluating multi-target tracking and person re-identification, they are collected by static cameras.
In comparison, MDOT can dynamically track the targets with moving drones (see Figure \ref{dataset}).
Note that we do not collect the lidar data because several drones with lidar are much expensive compared with visible light camera equipped drones and the accurate sensing distance of lidar is about 200m, which does not
show obvious advantage compared with visible light cameras.
\begin{figure}
	\centering
	\includegraphics[width=0.95\linewidth]{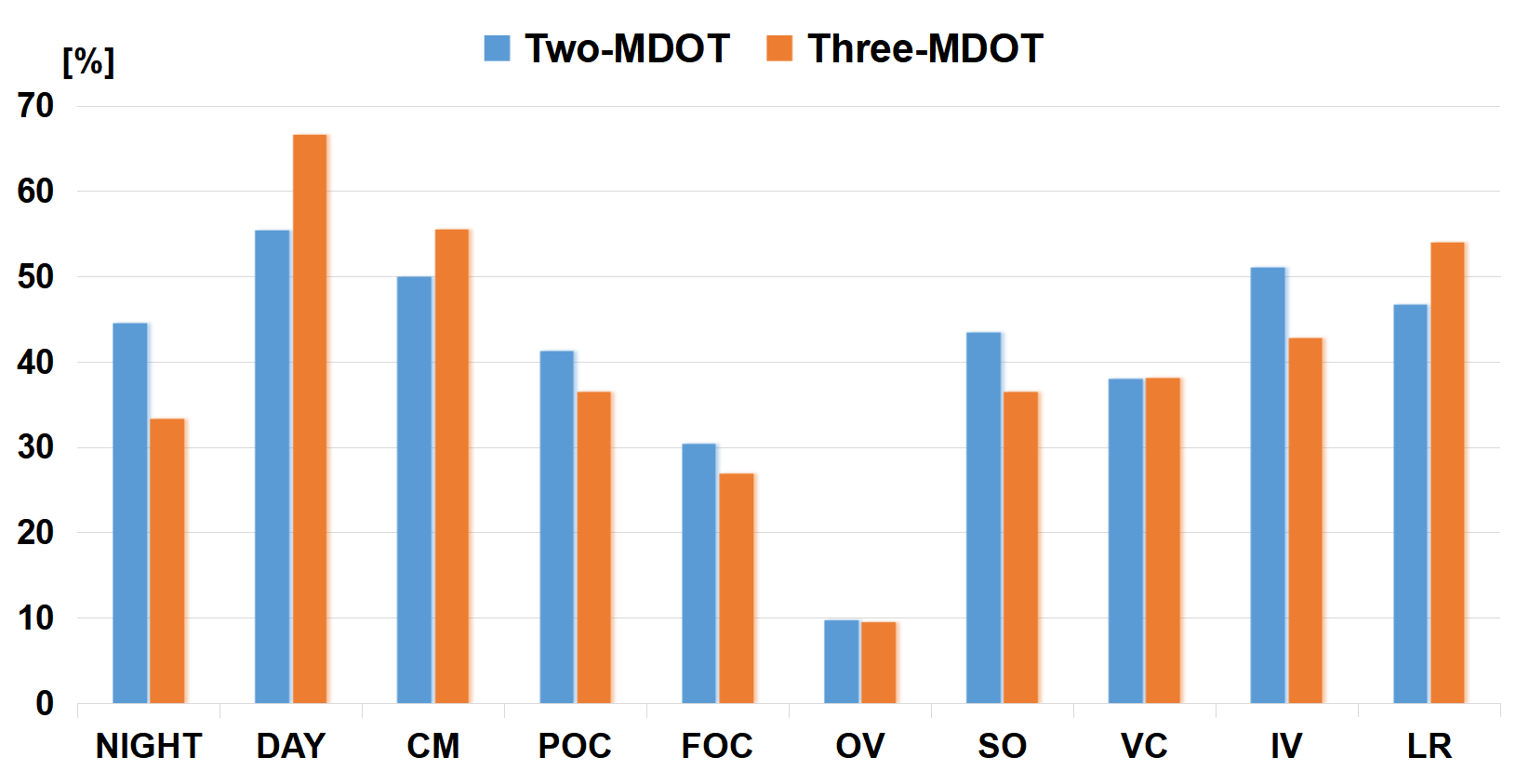}
	\caption{Statistics with respect to $10$ visual attributes.}
	\label{attribute}
\end{figure}

\subsection{Annotation}
For annotation, we collect the images with the size of $1280\times720$ and use the commonly used annotation tool VATIC \cite{springerlink:10.1007/s11263-012-0564-1} to annotate the location, occlusion and out\_of\_view information of targets. After that, LabelMe \cite{russell2008labelme} is used to refine and double-check the annotations frame-by-frame. Moreover, the targets in $155$ sequences are divided into $9$ categories, \ie, pedestrian, car, carriage, motor, bicycle, tricycle, truck, dog, bus, and the targets in each category are also diverse. Moreover, as shown in Table \ref{tab:attribute}, all the sequences are labeled by $10$ attributes, \ie, Daytime (DAY), Night (NIGHT), Camera Motion (CM), Partial Occlusion (POC), Full Occlusion (FOC), Out of View (OV), Similar Object (SO), Viewpoint Change (VC) and Illumination Variation(IV), Low Resolution (LR). The statistics with respect to $10$ attributes are summarized in Figure \ref{attribute}. Notice that CM, IV and LR occur in most sequences, which may significantly degrade the performance of trackers.
Similar to the setting of the classic single object tracking task, we manually annotate the tracking target in the first frame across different drones with respect to the same object.
\begin{table*}[t]
	\centering
	\caption{Descriptions of each attribute used in the MDOT dataset.}
	\label{tab:attribute}
	\footnotesize
	\setlength{\tabcolsep}{18.0pt}
	\begin{tabular}{c|l}
		\hline
		Attribute                             &Description \\
		\hline
		DAY                                   &Daytime: the sequence is taken during the daytime. \\
		NIGHT                                 &Night: the sequence is taken at night. \\
		CM                                    &Camera Motion: abrupt motion of the camera. \\
		POC                                   &Partial Occlusion: the target is partially occluded in the sequence. \\
		FOC                                   &Full Occlusion: the target is fully occluded in the sequence. \\
		OV                                    &Out Of View: some frames of the target leave the view. \\
		SO                                    &Similar Object: there are targets of similar shape or same type near the target.\\
		VC                                    &Viewpoint Change: viewpoint affects target appearance significantly. \\
		IV                                    &Illumination Variation: the illumination in the target region changes. \\
		LR                                    &Low Resolution: the frame number of tiny targets (pixels are less than $400$) is more than $50$.  \\
		\hline
	\end{tabular}
\end{table*}

\begin{figure*}[t]
	\centering
	\includegraphics[width=0.95\linewidth]{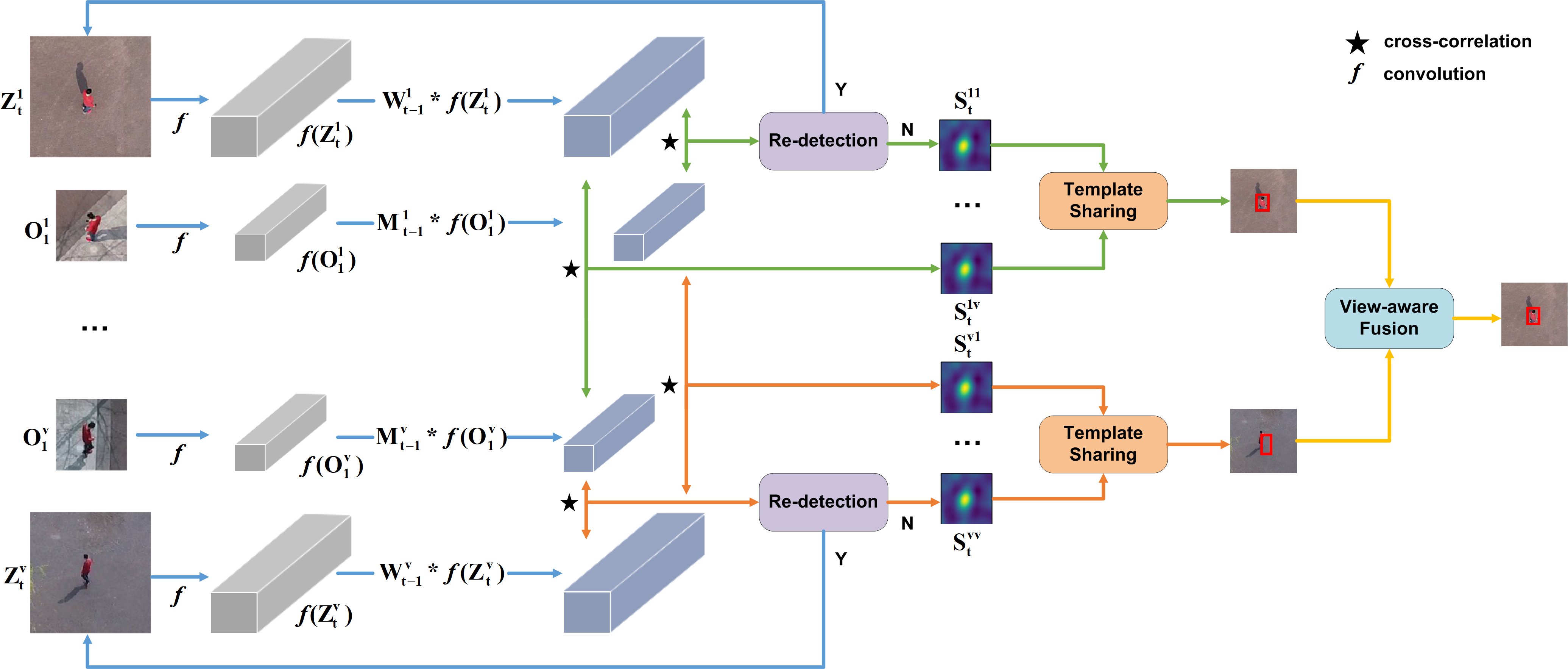}
	\caption{The architecture of agent sharing network (ASNet), which consists of re-detection, template sharing and view-aware fusion modules.}
	\label{framework}
\end{figure*}

\subsection{Evaluation Metrics}
Single object tracking is usually evaluated by success and precision plots \cite{wu2015object}. However, for multi-drone based tracking, the results of the algorithms should be evaluated upon the multi-drone fusion results.
To this end, we propose two new metrics for multi-drone single object tracking, \ie, automatic fusion score (AFS) and ideal fusion score (IFS).
\begin{itemize}
\item \textbf{Automatic fusion score} evaluates the performance of multi-drone tracker that fuses the tracking results using the online fusing strategy.
\begin{myDef}\label{afs}
	Let ${{\bf{h}}_i^v}$ and ${{\bf{y}}_i^v}$ be the tracking result (\ie, location, width and height of the box) and ground truth of the $i$-th frame and $v$-th drone. \text{AFS} is defined as
	\begin{equation}
	\text{AFS} =\frac{{\rm{1}}}{n} \sum\nolimits_{i = 1}^n {\sum\nolimits_{v = 1}^V {{w_v}s({\bf{h}}_i^v,{\bf{y}}_i^v)} },
	\end{equation}
	where $s(\cdot,\cdot)$ is a evaluation metric for single object tracking (\ie, success and precision scores) and $w_v$ is the weight for the $v$-th drone.
    $n$ and $V$ is the number of frames in a video clip and the number of drones, respectively.
	The value of $w_v$ should be zero or one. $w_v$ is automatically learned and online updated for each frame during the tracking process.
\end{myDef}

\item \textbf{Ideal fusion score} is the ideal fusion performance when we assume that the multi-drone system can accurately select the tracking results of the drone with better performance. It is defined to evaluate the extreme performance of a multi-drone tracking system, which can guide the design of a superior multi-drone tracker.
\begin{myDef}\label{IFS}
	Let ${{\bf{h}}_i^v}$ and ${{\bf{y}}_i^v}$ be the tracking result and ground truth of the $i$-th frame and $v$-th drone. IFS is defined as
	\begin{equation}
	\text{IFS} =\frac{{\rm{1}}}{n} \sum\nolimits_{i = 1}^n {\max (s({\bf{h}}_i^1,{\bf{y}}_i^1),\cdots,s({\bf{h}}_i^V,{\bf{y}}_i^V))}.
	\end{equation}
\end{myDef}

\end{itemize}

During the evaluation stage, OPE metrics (Success/Precision) are the traditional metrics for single object tracker.
Based on the OPE, AFS and IFS are proposed for multi-drone tracker.
Compared with AFS, IFS shows that there is still a gap from the upper bound of multi-drone fusion, which inspires us to design superior multi-drone tracker with more useful fusion strategies.

\section{Agent Sharing Network}
The key challenge of multi-drone tracking is how to share the inter-drone information and adaptively fuse the tracking results.
To deal with this challenge, each drone is considered as an agent and we propose an agent sharing network (ASNet) for multi-drone tracking, which can effectively exploit the inter-agent complementary information, see Figure \ref{framework}.

\subsection{Network Architecture}
Dynamic siamese network (DSiam)~\cite{guo2017learning} can enable effective online learning of target appearance variation and background suppression from previous frames. Therefore, we choose DSiam as the base tracker and develop the corresponding multi-drone tracker. A common tracker is trained for all drones in that all drones track the same target in the same scene. Therefore, there is no bias toward different drones. We focus on the online tracking process and design the agent sharing network from three aspects, \ie, template sharing, view-aware fusion and target re-detection.

Let ${\bf{O}}_1^v$ and ${\bf{Z}}_t^v$ denote the templates of the first frame and search regions of the $t$-th frame with respect to the $v$-th drone, respectively. By an embedding block $\bf{f}$, \eg, convolutional neural network (CNN), deep features can be extracted for both the templates and search regions, \ie, $f({{\bf{O}}_1^v})$, and $f({\bf{Z}}_t^v)$.
The key components of Dsiam are the target appearance variation transformation and background suppression transformation \cite{guo2017learning}. For ASNet, we need to determine the transformation for all drones. The target appearance variation transformation ${\bf{M}}^v_{t-1}$ with respect to the $v$-th drone is learned by
\begin{equation}\label{M}
{{\bf{M}}^v_{t - 1}} = \arg \min {\left\| {{\bf{M \otimes F}}_1^v - {\bf{F}}_{t - 1}^v} \right\|^2} + {\lambda _m}{\left\| {\bf{M}} \right\|^2},
\end{equation}
where ${\bf{F}}_1^v = f({\bf{O}}_1^v)$ and ${\bf{F}}_{t - 1}^v = f({\bf{O}}_{t - 1}^v)$. ${\bf{O}}_{t - 1}^v$ is the tracked target of the ${t-1}$-th frame for the $v$-th drone. $\otimes$ denotes the circular convolution, which can computed rapidly in the frequency domain \cite{guo2017learning}. ${{\bf{M}}^v_{t-1}} $ can capture the target variation under temporal smoothness assumption and therefore contributes greatly to online learning. Similarly, the background suppression transformation ${\bf{W}}^v_{t-1}$ can be learned.
\begin{equation}
\label{W}
{{\bf{W}}_{t - 1}} = \arg \min {\left\| {{\bf{W \otimes F}}_{G_{t - 1}^v}^v - {\bf{F}}_{\tilde G_{t - 1}^v}^v} \right\|^2} + {\lambda _w}{\left\| {\bf{W}} \right\|^2},
\end{equation}
where ${{G_{t-1}^v}}$ is the region centering at the target with the same size of ${\bf{Z}}_{t-1}^v$. ${\tilde G_{t - 1}^v}$ is obtained by multiplying ${{G_{t-1}^v}}$ with a Gaussian weight map. ${{\bf{W}}_{t - 1}}$ can suppress the background information and therefore induces superior tracking performance. More details about the solution of ${{\bf{M}}^v_{t - 1}}$ and ${{\bf{W}}_{t-1}}$ can be found in \cite{guo2017learning}.

Compared with visual tracking using single drone, ASNet shares the templates of all drones, and obtains the response maps corresponding to the templates of multiple drones. As the reliability of templates of multiple drones is different, we adaptively fuse the response maps of multiple-templates in a self-supervised manner. Finally, the tracking results of multiple drones can be adaptively fused by tracking scores.


\subsection{Self-supervised Template Sharing}
As the appearance of the target may vary greatly, the templates of all drones can be shared to improve the tracker robustness of single drone. The response map ${\bf{S}}_t^{kv}$ of the tracker on the $k$-th drone, corresponding to the template of the $v$-th drone is calculated as
\begin{equation}
\label{responsemap}
{\bf{S}}_t^{kv} = \text{corr}({\bf{M}}_{t - 1}^v \otimes f({\bf{O}}_1^v),{\bf{W}}_{t - 1}^k  \otimes f({\bf{Z}}_t^k)),
\end{equation}
where $\text{corr}(\bf{A},\bf{B})$ is the correlation operation, which can be considered as a convolution operation on $\bf{B}$ with $\bf{A}$ as the convolution filter. For the $k$-th drone, we obtain a set of response maps, ${\bf{S}}_t^{k1}, {\bf{S}}_t^{k2},\cdots, {\bf{S}}_t^{kV}$. To fuse the response maps, we propose a self-supervised fusion strategy. Specifically, we use the tracking results of the previous $t-1$ frames as the supervised information to guide the weights learning of $V$ templates.
Let ${\bf{O}}_t^{kv}$ denote the tracked target of the tracker on the $k$-th drone using the template of the $v$-th drone.
The fusion weights ${{\bf{u}}_t^k}$ can be learned by
\begin{equation}
\label{fusionweights }
{\min _{\bf{u}}}{\left\| {{\bf{D}}_t^k{\bf{u}}_t^k - {\bf{M}}_{t - 1}^k \otimes f({\bf{O}}_1^k)} \right\|^2} + {\lambda _{\bf{u}}}{\left\| {{\bf{u}}_t^k} \right\|^2},
\end{equation}
where ${\bf{D}}_t^k = \left[ {f({\bf{O}}_t^{k1}),f({\bf{O}}_t^{k2}),\cdots,f({\bf{O}}_t^{kV})} \right].$
The weights reflect the correlation between tracked target and target template of the $(t-1)$-th frame. Given the learned weights and response maps, we can obtain the fused response map for the tracker on the $k$-th drone, \ie,
\begin{equation}\label{fusemap}
{\bf{S}}_t^k = \sum\nolimits_{v = 1}^n {u_t^{kv}{\bf{S}}_t^{kv}}.
\end{equation}
For a multi-drone tracking system with $V$ drones, we can obtain $V$ fused response maps in total, \ie, ${\bf{S}}_t^1, {\bf{S}}_t^2,\cdots, {\bf{S}}_t^V$.

\subsection{View-aware Fusion}
To generate the final results on the multi-drone tracking, we use the auto view-aware fusion scheme when we obtain the $V$ fused response maps. For the $k$-th drone's response map, we search the maximum value $g({{\bf{s}}_t^k})$ in the response map, and obtain its respective location ${{\bf{p}}_t^k}$. $g({{\bf{s}}_t^k})$ is defined as tracking score with respect to the $t$-th frame on the $k$-th tracker. Then, we can obtain the index of the best response map ${{\bf{b}}_t}$ by
\begin{equation}
\label{View-aware}
{{\bf{b}}_t} = \arg \max\big( g({{\bf{s}}_t^1}),g({{\bf{s}}_t^2}),\cdots,g({{\bf{s}}_t^v}) \big).
\end{equation}
The respective location ${{\bf{p}}_t^{{\bf{b}}_t}}$ in the ${{\bf{b}}_t}$-th drone is the position of the target.
The weight $w_v$ of the drone tracker with the best response map obtained from Eq. \ref{View-aware} is set as one and the rest as zero.

\subsection{Target Re-detection}
As camera motion often occurs in the drone based tracking, the target location may vary dramatically in successive frames. To solve the problem, we use the target re-detection strategy based on the past and current frames. For the $t$-th frame, let ${{\bf{l}}_t^q}$ denote a set of scores for the past $q$ frames. $\mu_t^n$ and $\sigma_t^q$ denote mean and standard deviation of ${{\bf{l}}_t^q}$. Inspired by the peak to sidelobe ration in MOSSE \cite{bolme2010visual}, the threshold for target re-detection is defined as
\begin{equation}
\label{re-detection}
\omega_t^q = \mu_t^q - \lambda \cdot \sigma_t^q,
\end{equation}
where $\lambda$ is a pre-set parameter. The target may be lost when the score is less than $\omega_t^q$ or the tracking score is the threshold $T_{score}$. If so, we use the local-to-global strategy to expand the search region step by step to re-detect the target \cite{zhu2018distractor}. After using the proposed re-detection strategy, the tracking performance is greatly improved.

\section{Experiments}
We evaluate our method compared with $20$ recent state-of-the-art single object tracking algorithms on the machine with a E5-2620 v3 CPU and a NVIDIA TITAN Xp GPU.
Note that the existing multi-camera tracking methods, are specially designed for multi-object tracking and therefore cannot apply to multi-drone single object tracking directly.
The source codes for other algorithms are from the authors.

\subsection{Overall Performance}
We report the success and precision scores, tracking speed and the references of each algorithm in Table \ref{Two-MDOT} and Table \ref{Three-MDOT}.
The tracking performance on each drone are reported using the baseline single object trackers.
Besides, the overall performance of the baseline trackers is given by calculating success and precision scores of all drones together.
Note that for the proposed ASNet, we report success and precision plots using AFS in Definition \ref{afs} in terms of multi-drone tracking.
As shown in Table \ref{Two-MDOT} and Table \ref{Three-MDOT}, GFSDCF achieves the best precision score of $59.5$ on Two-MDOT and $65.6$ on Three-MDOT.
Following the GFSDCF tracker, other correlation filter trackers also obtain the great performance in precision score and success score, \eg, ECO, STRCF, CSRDCF.
Besides, due to extensive offline training, siamese tracking approaches DSiam, SiamRPN++ and SiameseFC show the top performance as well.
Our proposed ASNet significantly outperforms the baseline trackers on all sub-datasets.
Compared with the best baseline tracker, the precision scores are improved by $14.8$ on Two-MDOT and $12.6$ on Three-MDOT, respectively.
The results show that compared with tracking using only one drone, multi-drone tracking using our proposed ASNet greatly boost the tracking performance, which validates the
necessity and effectiveness of multi-drone tracking.
The significant improvement comes from the fusion of complementary information across multi-agents in case of great appearance variations.

\begin{figure}[t]
	\centering
	\includegraphics[width=1.0\linewidth]{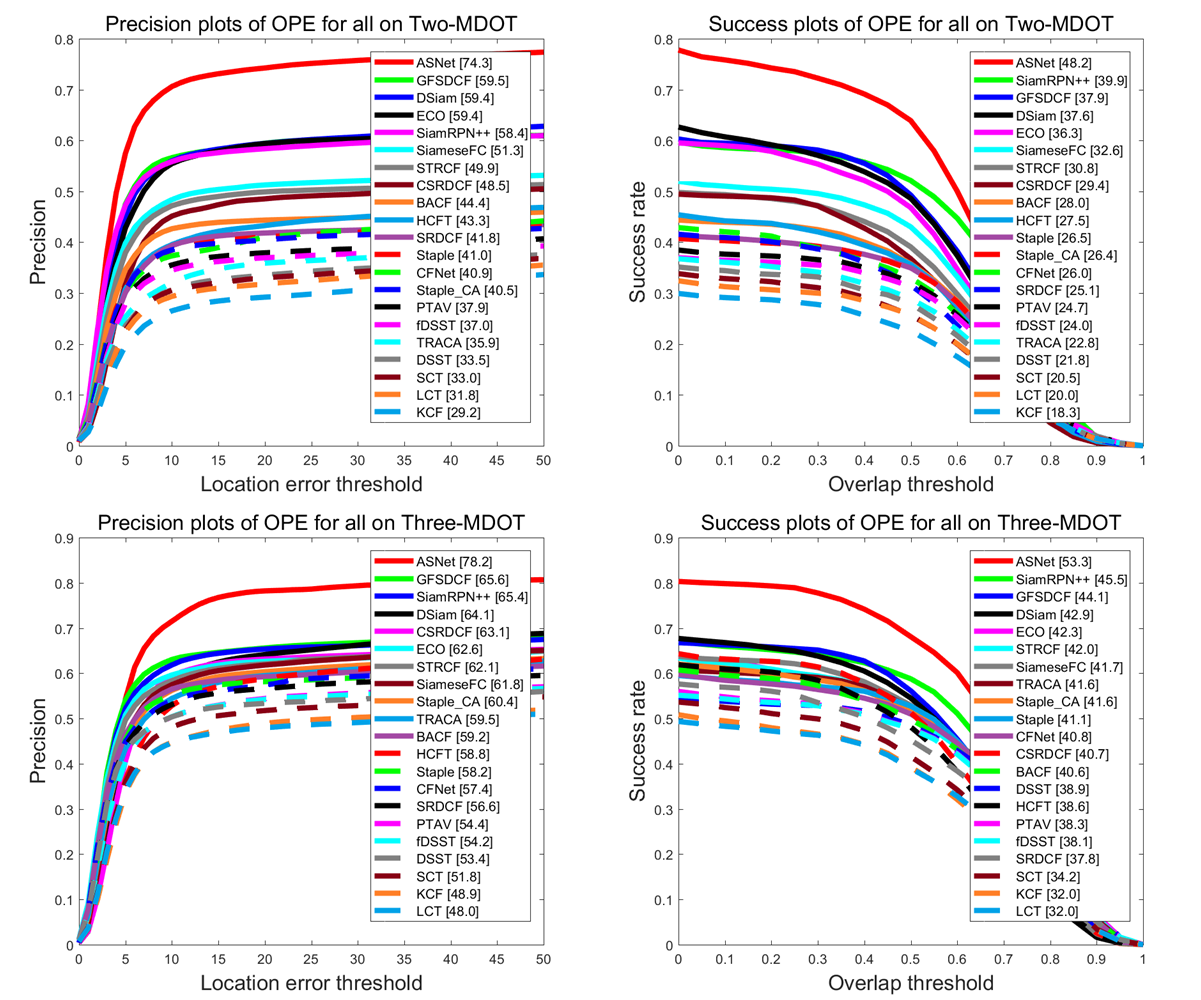}
	\caption{Success and precision plots on the MDOT test set. Note that our proposed ASNet uses the AFS metric.}
	\label{overall}
\end{figure}

To further analyze the performance, we report the overall performance of the proposed ASNet and $20$ compared state-of-the-art trackers in Figure \ref{overall}.
Notably, the success and precision plots of ASNet are drawn based on the AFS metric defined in our paper.
ASNet chieves the best performance on the proposed dataset, \ie, $48.2$ success score on the Two-MDOT subset and $53.3$ success score on the Three-MDOT subset.
This is because of the fusion of complementary information across multi-agents in case of great appearance variations in our framework.
We can conclude that SiamRPN++ \cite{li2018siamrpn++} and GFSDCF \cite{xu2019joint} rank the second and third place compared with other methods in terms of success score, respectively.
Following the above three trackers, the siamese network based tracker DSiam \cite{guo2017learning} and correlation filters based tracker ECO \cite{danelljan2017eco} obtain slight inferior performance in both success and precision scores.

\begin{table*}[t]
	\centering
	\footnotesize
	\caption{Comparisons on Two-MDOT.}
	\label{Two-MDOT}
	\setlength{\tabcolsep}{10.0pt}
	\begin{tabular}{c|c|c|c|c|c|c|c|c}
		\hline
		\multirow{2}{*}{Algorithms}  & \multicolumn{2}{c|}{Drone1}     & \multicolumn{2}{c|}{Drone2}   & \multicolumn{2}{c|}{overall}   & \multirow{2}{*}{Speed}  & \multirow{2}{*}{Reference} \\
		\cline{2-7}
		&Success  &Precision                   &Success  &Precision    &Success  &Precision  & &\\
		\hline
		DSST       &24.4   &40.1   &19.3  &27.0    &21.8    &33.5    &76.2  &BMVC'14~\cite{danelljan2014accurate} \\
		SRDCF      &23.4   &41.4   &26.8  &42.3    &25.1    &41.8    &14.7  &ICCV'15~\cite{danelljan2015learning} \\
		KCF        &18.6   &31.5   &17.9  &26.9    &18.3    &29.2    &441.6  &T-PAMI'15~\cite{henriques2015high}   \\
		LCT        &19.3   &32.6   &20.7  &31.1    &20.0    &31.8    &33.7  &CVPR'15~\cite{ma2015long}            \\
		HCFT       &29.1   &46.8   &25.9  &39.7    &27.5    &43.3    &1.4  &ICCV'15~\cite{ma2015hierarchical}    \\
		Staple     &29.0   &46.4   &24.1  &35.7    &26.5    &41.0    &61.2  &CVPR'16~\cite{Bertinetto_2016_CVPR}  \\
		SCT        &19.5   &33.2   &21.5  &32.8    &20.5    &33.0    &38.2  &CVPR'16~\cite{choi2016visual}        \\
		SiameseFC  &33.6   &53.6   &31.6  &48.9    &32.6    &51.3    &19.5  &ECCVW'16~\cite{bertinetto2016fully}  \\
		fDSST      &24.5   &40.0   &23.5  &33.9    &24.0    &37.0    &112.2  &T-PAMI'17~\cite{danelljan2017discriminative}  \\
		Staple\_CA &27.8   &44.6   &25.0  &36.3    &26.4    &40.5    &36.5  &CVPR'17~\cite{mueller2017context}    \\
		BACF       &29.5   &49.6   &26.4  &39.1    &28.0    &44.4    &35.5  &ICCV'17~\cite{kiani2017learning}     \\
		CSRDCF     &31.9   &54.6   &26.9  &42.4    &29.4    &48.5    &8.4  &CVPR'17~\cite{lukezic2017discriminative}  \\		
		ECO        &37.0   &62.2   &35.6  &\textbf{56.6}&36.3    &59.4    &13.7  &CVPR'17~\cite{danelljan2017eco}  \\		
		DSiam      &39.0   &63.7   &36.1  &55.2    &37.6    &59.4    &18.9  &ICCV'17~\cite{guo2017learning}       \\		
		PTAV       &25.3   &41.3   &24.1  &34.6    &24.7    &37.9    &13.3  &ICCV'17~\cite{fan2017parallel}       \\		
		CFNet      &28.0   &45.5   &24.1  &36.2    &26.0    &40.9    &17.7  &CVPR'17~\cite{valmadre2017end}       \\		
		TRACA      &23.1   &38.5   &22.5  &33.3    &22.8    &35.9    &58.1  &CVPR'18~\cite{choi2018context}       \\		
		STRCF      &31.4   &52.2   &30.3  &47.6    &30.8    &49.9    &22.3  &CVPR'18~\cite{li2018learning}        \\		
		SiamRPN++  &\textbf{42.2}&62.2&\textbf{37.5}&54.6 &\textbf{39.9}&58.4 &39.6  &CVPR'19~\cite{li2018siamrpn++}\\		
		GFSDCF     &41.7&\textbf{67.6}&34.2& 51.4 &37.9&\textbf{59.5} &5.5  &ICCV'19~\cite{xu2019joint} \\		
		\hline
		ASNet      &\textbf{-}&\textbf{-}&\textbf{-}&\textbf{-} &\textbf{48.2}&\textbf{74.3} &18.5  &ours\\
		\hline
	\end{tabular}
\end{table*}

\begin{table*}[t]
	\centering
	\footnotesize
	\caption{Comparisons on Three-MDOT.}
	\label{Three-MDOT}
	\setlength{\tabcolsep}{7.0pt}
	\begin{tabular}{c|c|c|c|c|c|c|c|c|c|c}
		\hline
		\multirow{2}{*}{Algorithms}  & \multicolumn{2}{c|}{Drone1}     & \multicolumn{2}{c|}{Drone2}   & \multicolumn{2}{c|}{Drone3} & \multicolumn{2}{c|}{overall}   & \multirow{2}{*}{Speed}  & \multirow{2}{*}{Reference} \\
		\cline{2-9}
		&Success  &Precision     &Success  &Precision &Success  &Precision    &Success  &Precision  & &\\
		\hline
		DSST       &37.0 &49.8 &38.6 &52.5 &41.0 &57.8 &38.9 &53.4 &68.1  &BMVC'14~\cite{danelljan2014accurate} \\
		SRDCF      &34.3 &49.4 &38.5 &60.7 &40.4 &59.8 &37.8 &56.6 &12.9  &ICCV'15~\cite{danelljan2015learning} \\
		KCF        &31.6 &46.5 &30.5 &50.4 &33.9 &50.0 &32.0 &48.9 &411.3 &T-PAMI'15~\cite{henriques2015high}   \\
		LCT        &30.5 &43.7 &33.1 &52.5 &32.4 &47.6 &32.0 &48.0 &29.9  &CVPR'15~\cite{ma2015long}            \\
		HCFT       &36.2 &55.4 &37.6 &57.7 &41.9 &63.5 &38.6 &58.8 &1.79  &ICCV'15~\cite{ma2015hierarchical}    \\
		Staple     &39.4 &55.1 &42.3 &60.9 &41.7 &58.4 &41.1 &58.2 &59.5  &CVPR'16~\cite{Bertinetto_2016_CVPR}  \\
		SCT        &34.9 &52.8 &30.9 &48.0 &36.8 &54.5 &34.2 &51.8 &30.9  &CVPR'16~\cite{choi2016visual}        \\
		SiameseFC  &41.9 &63.6 &40.8 &59.7 &42.5 &62.1 &41.7 &61.8 & 18.3 &ECCVW'16~\cite{bertinetto2016fully}  \\
		fDSST      &33.1 &46.0 &41.5 &60.5 &39.7 &57.8 &38.1 &54.2 &96.8  &T-PAMI'17~\cite{danelljan2017discriminative}  \\
		Staple\_CA &40.5 &58.2 &41.0 &61.8 &43.3 &61.1 &41.6 &60.4 &35.8  &CVPR'17~\cite{mueller2017context}    \\
		BACF       &39.1 &58.0 &42.4 &62.3 &40.4 &57.4 &40.6 &59.2 &33.1  &ICCV'17~\cite{kiani2017learning}     \\
		CSRDCF     &39.4 &62.4 &41.0 &63.0 &41.9 &64.0 &40.7 &63.1 &8.5   &CVPR'17~\cite{lukezic2017discriminative}  \\		
		ECO        &41.6 &60.8 &42.1 &62.6 &43.1 &64.5 &42.3 &62.6 &12.5  &CVPR'17~\cite{danelljan2017eco}  \\		
		DSiam      &42.3 &63.5 &42.7 &62.8 &43.6 &\textbf{66.1} &42.9 &64.1 &10.4 &ICCV'17~\cite{guo2017learning}    \\		
		PTAV       &32.7 &45.2 &41.6 &60.3 &40.6 &57.7 &38.3 &54.4 &11.8  &ICCV'17~\cite{fan2017parallel}  \\		
		CFNet      &40.6 &57.2 &40.3 &55.3 &41.5 &59.7 &40.8 &57.4 &18.7  &CVPR'17~\cite{valmadre2017end}   \\		
		TRACA      &41.8 &59.2 &40.6 &58.8 &42.6 &60.4 &41.6 &63.1 &49.1  &CVPR'18~\cite{choi2018context}  \\		
		STRCF      &40.3 &59.6 &42.0 &60.7 &43.7 &65.9 &42.0 &46.1 &21.0  &CVPR'18~\cite{li2018learning}   \\		
		SiamRPN++  &\textbf{44.5} &65.5 &\textbf{46.6} &66.8 &\textbf{45.3} &63.8 &\textbf{45.5} &65.4 &39.6  &CVPR'19~\cite{li2018siamrpn++}\\		
		GFSDCF     &42.8 &\textbf{66.5} &45.5 &\textbf{67.3} &44.1 &64.0 &44.1 &\textbf{65.6} &5.4  &ICCV'19~\cite{xu2019joint} \\		
		\hline
		ASNet      &\textbf{-} &\textbf{-}&\textbf{-} &\textbf{-}&\textbf{-} &\textbf{-}&\textbf{53.3} &\textbf{78.2} &18.5  &ours\\
		\hline
	\end{tabular}
\end{table*}

\begin{table*}
	\centering
	\footnotesize
	\caption{Effectiveness of the three components, \ie, re-detection, template sharing, view-aware fusion, in the proposed method.}
	\label{tab:ablation}
	\setlength{\tabcolsep}{4.5pt}
	\begin{tabular}{l|c|c|c|c|c|c|c}
		\hline
		\multirow{2}{*}{Algorithms} & \multirow{2}{*}{Re-detection} & \multirow{2}{*}{Template Sharing} & \multirow{2}{*}{View-aware Fusion} & \multicolumn{2}{c|}{Two-MDOT}     & \multicolumn{2}{c}{Three-MDOT}  \\
		\cline{5-8}
		&                               &                                   &                                    &Success         &Precision           &Success         &Precision      \\
		\hline
		(1) DSiam                   &                               &                                   &                 &37.6           &59.4             &42.9    &64.1     \\
		\hline
		(2) Re-detection            &  $\surd$                      &                                   &                 &39.1           &62.2             &44.1    &66.2     \\
		(3) Template Sharing        &                               &  $\surd$                          &                 &38.2           &60.3             &43.5    &65.4     \\
		(4) ASNet w/o VF            &  $\surd$                      &  $\surd$                          &                 &39.6           &62.9             &44.7    &67.0     \\
		\hline
		(5) View-aware Fusion       &                               &                                   &  $\surd$        &46.3           &71.6             &52.0    &75.2     \\
		(6) ASNet w/o RD            &                               &  $\surd$                          &  $\surd$        &47.0           &72.7             &52.1    &76.0     \\
		(7) ASNet w/o TS            &  $\surd$                      &                                   &  $\surd$        &47.6           &73.4             &52.7    &76.6     \\
		\hline
		(8) ASNet                   &  $\surd$                      &  $\surd$                          &  $\surd$        &\textbf{48.2}  &\textbf{74.3}    &\textbf{53.3}   &\textbf{78.2} \\
		\hline
	\end{tabular}	
\end{table*}

\begin{figure}[t]
	\centering
	\includegraphics[width=\linewidth]{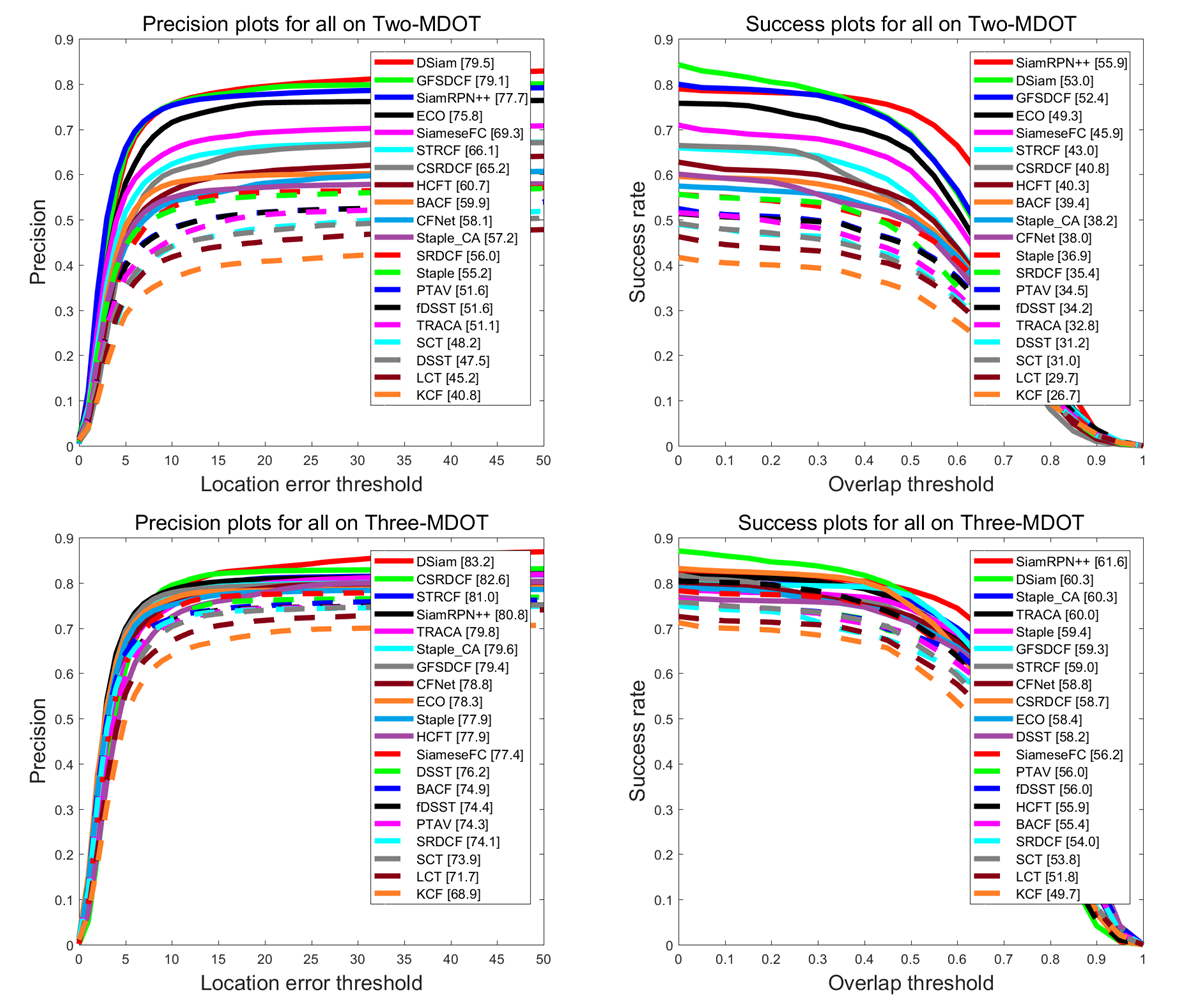}
	\caption{Success and precision plots of the existing methods with IFS metric on MDOT test set.}
	\label{fig:IFS}
\end{figure}

\subsection{Attribute-based Performance}
\begin{figure*}[t]
	\centering
	\includegraphics[width=1\linewidth]{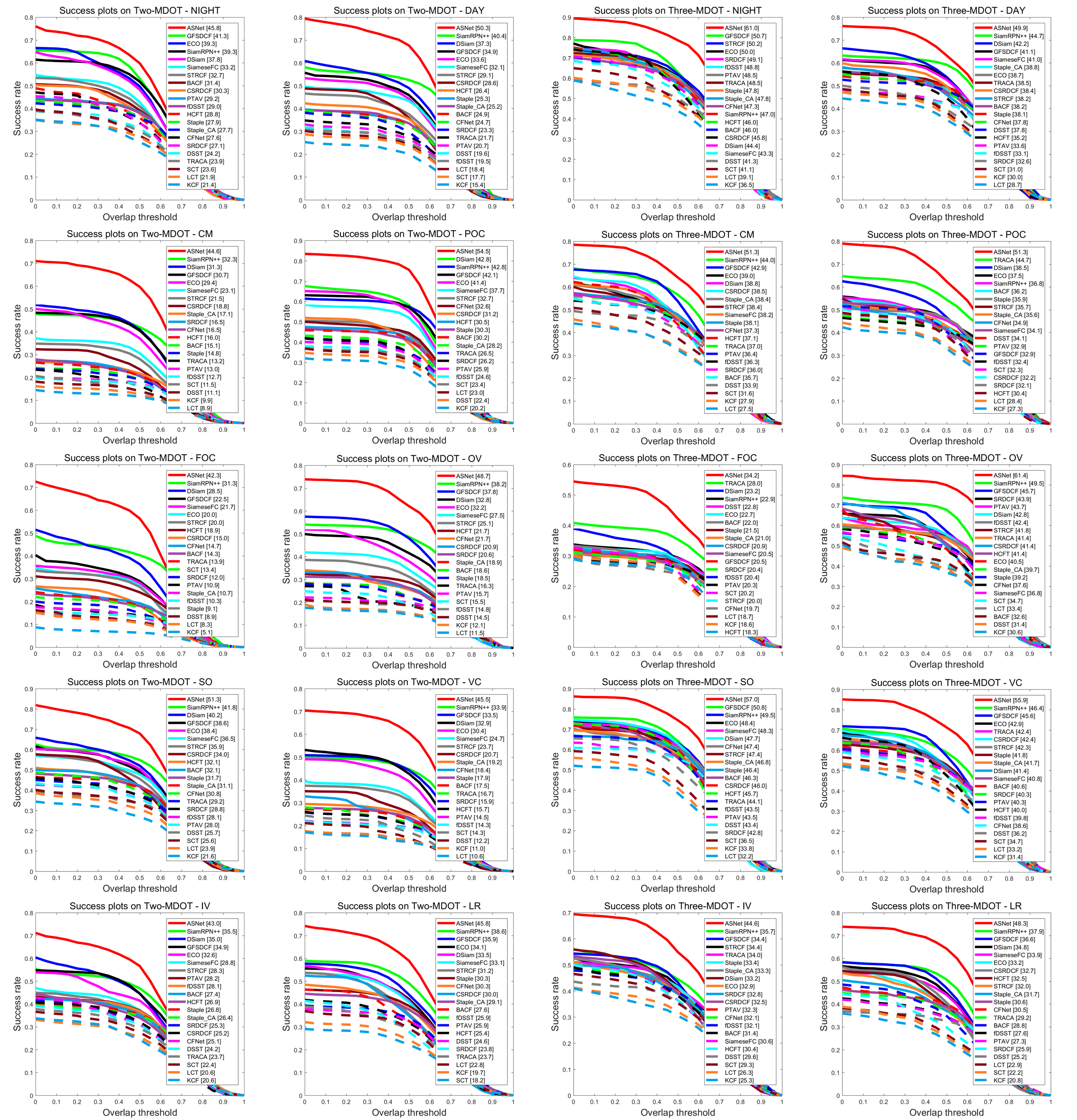}
	\caption{Success plots of state-of-the-art trackers on MDOT test set in terms of each attribute.}
	\label{attribute_result}
\end{figure*}

To further analyze the performance, we report the success scores of algorithms over $10$ attributes in Figure~\ref{attribute_result}.
It can be concluded that the performances on attributes CM, FOC, VC and IV are inferior than that on other attributes. This is maybe because the target appearance is heavily changed in these situations. We can observe that the performance of DSiam on attributes CM, SO, VC are far ahead of other methods. For other attributes, DSiam achieves the best performance on the test set, while the gap between DSiam and the best competitor ECO on the other UAV based datasets is small. Moreover, our proposed tracker ASNet achieves the best performance on all attributes. There is a significant gap between ASNet and other compared trackers, which owes to the discriminative appearance information from template sharing and re-detection strategy.

\begin{figure*}[t]
	\centering
	\includegraphics[width=1\linewidth]{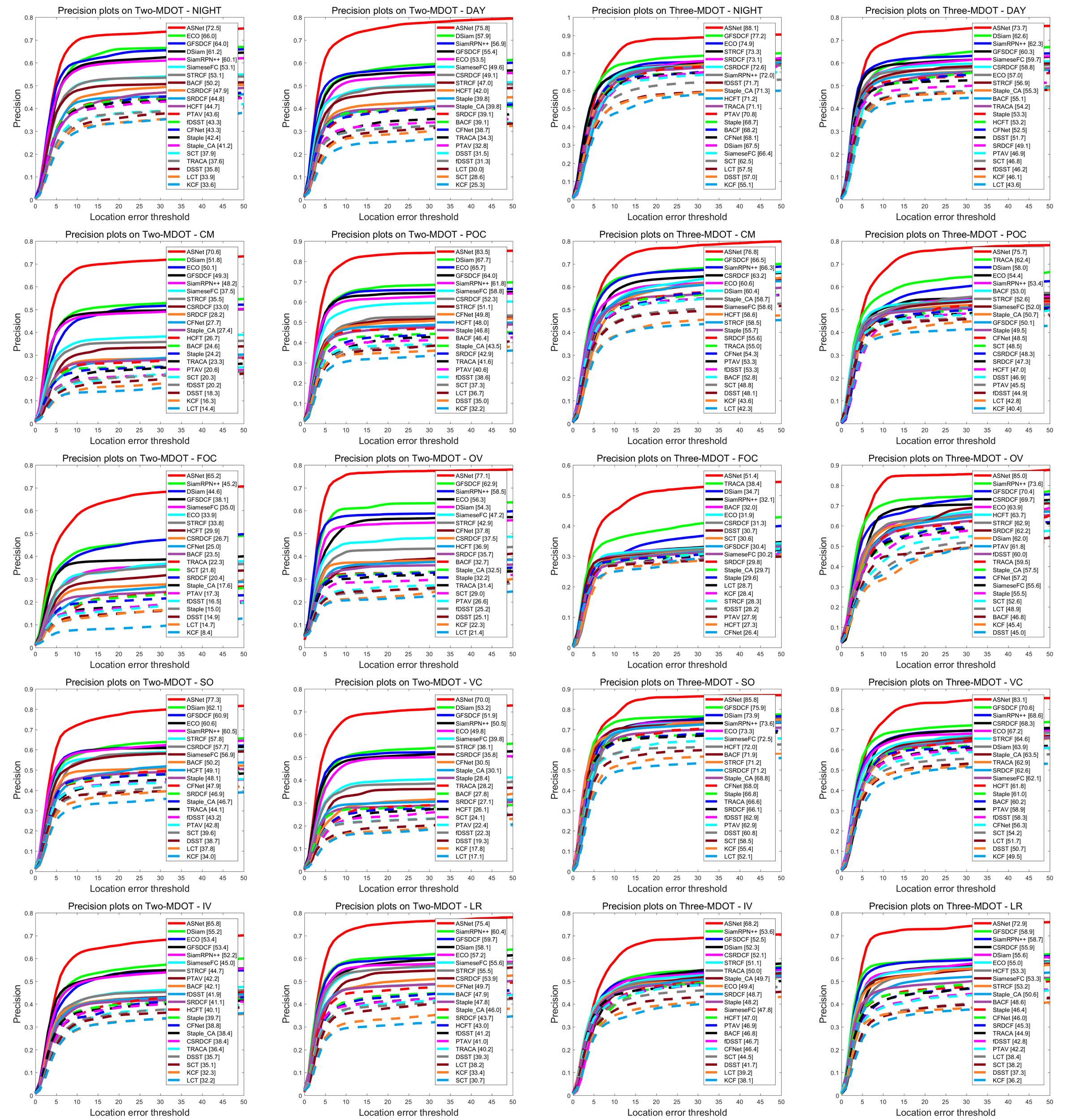}
	\caption{Precision plots of state-of-the-art trackers on MDOT test set in terms of each attribute.}
	\label{attribute_result_precision}
\end{figure*}

Figure~\ref{attribute_result_precision} shows the precision plots of compared tracking algorithms over $10$ challenging visual attributes. It can be concluded that the performances on attributes CM, FOC, VC and IV are inferior than that on other attributes. This is maybe because the target appearance is heavily changed in these situations. Moreover, our proposed tracker ASNet achieves the best performance on all attributes. There is a significant gap between ASNet and other compared trackers, which is attributed to the discriminative appearance representation from template sharing and re-detection strategy in our method. Following our method, DSiam, SiamRPN++ and GFSDCF achieve good performance in most attributes, much better than the remain compared tracking methods.

\subsection{Ideal Fusion Score}
To investigate the ideal performance of a multi-drone tracking system, as shown in Figure \ref{fig:IFS}, we report the success and precision plots of the existing trackers with IFS metric on MDOT test set.
Note that there is a big difference in the ideal fusion performance of different trackers.
Similar to ensemble learning techniques, IFS is up to the performance of the base tracker on each drone.
If the tracking results of the baseline tracker, \eg, DSiam, can be ideally fused with the precision of $79.5$ on Two-MDOT and $83.2$ on Three-MDOT, respectively.
For any single object tracker, IFS can be used to guide the design of a multi-drone tracker based on the base tracker.
As the tracking mechanism of the base trackers are different, more generalized or base tracker-specific fusion strategies are expected to be designed by the research community to boost the performance of multi-drone tracking.

\subsection{Ablation Study}
As shown in Table \ref{tab:ablation}, we analyze the importance of each component in ASNet on our MDOT dataset, \ie, re-detection, template sharing and view-aware fusion. Specifically, we use DSiam as the base tracker of ASNet and add re-detection, template sharing, view-aware fusion.

\noindent\textbf{Re-detection.}
We first add the re-detection component in the DSiam tracker. In this module, the hyperparameter $\lambda$ is set to $2$ in Two-MDOT and $1.25$ in Three-MDOT. As shown in Table \ref{tab:ablation} (2), the re-detection module can improve precision score of $2.8/2.1$ and success score of $1.5/1.2$ on Two-MDOT and Three-MDOT respectively. It indicates that the re-detection module can decrease the possibility of target drifting especially for long-term tracking.

\noindent\textbf{Template Sharing.}
Table \ref{tab:ablation} (3) shows the results of template sharing based on the DSiam tracker. It brings additional improvements for both precision score ($0.9/1.3$) and success score ($0.6/0.6$). When we combine target re-detection and template sharing, the precision score and success score are further improved.

\noindent\textbf{View-aware Fusion.}
As presented in Table \ref{tab:ablation} (5), if we only apply the view-aware fusion in the DSiam tracker, the performance is greatly improved. Then we take the view-aware fusion strategy into account, consistent improvements can be achieved on both drones, as shown in Table \ref{tab:ablation} (6, 7). Finally, we add the re-detection module, template sharing design and view-aware fusion scheme to the DSiam tracker (ASNet). Thus we can observe considerable improvement in precision score ($14.9/14.1$) and success score ($10.6/10.4$).

\subsection{Discussions}
As there exists exchange of information across drones, we need to take the impact of synchronization and latency into account.
Similar to the setting of multi-camera tasks, we assume that videos of multiple drones are synchronous by starting tracking across drones simultaneously.
Re-detection is conducted on each drone separately while view-aware fusion only needs a tracking score value per drone.
Hence, the only component of ASNet, \ie, template Sharing, is affected by communication latency.
For ASNet, even if the template sharing strategy is not adopted, we can still get a superior performance, as shown in Table \ref{tab:ablation} (7).
In real-world applications, with the development of communication network technology, if the communication latency can be ignored, we can exploit the sharing of vision information across drones more effectively.
\section{Conclusions}
In this paper, we present a new multi-drone single-object tracking (MDOT) benchmark dataset for the object tracking community.
MDOT is an unique platform for developing drone based tracking algorithms and multi-drone tracking systems. Moreover, Two evaluation metrics, \ie, adaptive fusion score (AFS) and ideal fusion score (IFS) are proposed for multi-drone single object tracking. To exploit the complementary information across drones, an agent sharing network (ASNet) is proposed by sharing inter-drone templates, fusing multi-drone tracking results and re-detecting the targets. Extensive experiments on MDOT show that ASNet outperforms the state-of-the-art single object trackers, which validates the effectiveness of multi-drone tracking.


\bibliographystyle{IEEEtran}
\bibliography{reference}

\end{document}